\documentclass[10pt, conference]{IEEEtran}
\IEEEoverridecommandlockouts

\usepackage[noadjust]{cite}
\usepackage{amsmath,amssymb,amsfonts}
\usepackage{graphicx}
\graphicspath{{./figures/}{./}}
\usepackage{textcomp}
\usepackage[capitalise]{cleveref}
\usepackage{booktabs}

\begin{document}

\title{Reinforcement Learning for Scalable Logic Optimization with Graph Neural Networks}

\author{\IEEEauthorblockN{Xavier Timoneda, Lukas Cavigelli}\IEEEauthorblockA{Huawei Technologies, Zurich Research Center, Switzerland}\thanks{This preprint has been accepted for publication at DAC'21. © 2021 IEEE. Personal use of this material is permitted. Permission from IEEE must be obtained for all other uses, in any current or future media, including reprinting/republishing this material for advertising or promotional purposes, creating new collective works, for resale or redistribution to servers or lists, or reuse of any copyrighted component of this work in other works.}}

\maketitle

\begin{abstract}
Logic optimization is an NP-hard problem commonly approached through hand-engineered heuristics. We propose to combine graph convolutional networks with reinforcement learning and a novel, scalable node embedding method to learn which local transforms should be applied to the logic graph. We show that this method achieves a similar size reduction as ABC on smaller circuits and outperforms it by $1.5$--$1.75\times$ on larger random graphs. 

\end{abstract}

\section{Introduction \& Related Work} 
\label{sec:introduction} 

Logic synthesis is a crucial optimization step in the flow from the RTL description to the final GDS data. Minimizing the gate count of a logic circuit, even without timing constraints, is known to be NP-hard. With today's available compute power, logic functions with only a few inputs have come within reach of optimal synthesis (e.g., using SAT solvers). The synthesis of larger circuits, however, remains in the hand of algorithms relying on heuristics. 

Deep learning has revolutionized the field of computer vision and many others by replacing engineered heuristics with multi-layered learned heuristics, leading to dramatic leaps in the quality of results (QoR). In the following, we investigate how the game of logic optimization can be taught to graph convolutional neural networks (GCNs) \cite{Kipf2017} operating on a graph representation of logic circuits using reinforcement learning (RL). 

Two previous works have made first steps in this direction. DRiLLS \cite{Hosny2020} has applied RL at a coarse-grained level to select which commands/heuristics of UC Berkeley's ABC logic synthesis tool \cite{Epfl} should be applied, showing clear improvements in their QoR. Haaswijk \textit{et al.} \cite{Haaswijk2018} have investigated the direct application of RL to GCNs for logic optimization based on majority-inverter graphs (MIGs), showing that they can achieve optimality for small logic functions by comparing the results against synthesis based on SAT solvers. This method has the key limitations that 
1) its embedding method does not scale to larger circuits, requiring a larger embedding dimension as the circuit size increases, 
2) it has to be trained on a graph with an identical number of nodes as the one to be optimized, and 
3) it only applies a single action at one node of the graph, resulting in $\mathcal{O}(N^2)$ compute time. 

In this work, we combine RL with GCNs based on MIG logic graphs and focus on overcoming these limitations of \cite{Haaswijk2018}. Specifically, we contribute 
1) a novel embedding method that is independent of the graph size while demonstrating its capability to achieve a high QoR; 
2) a compact RL-based framework resulting in a GCN with only a few trainable parameters that enables inference on arbitrarily large graphs regardless of the graph size used during training; 
3) a GCN that learns an optimized set of \textit{atomic} actions for each node, applying local modifications at many nodes within the graph with each inference step; 
4) an evaluation of our method that includes 500-node random logic circuits.

\section{Proposed Method} 
\label{sec:systemdesign}

\begin{figure}
\centering
\includegraphics[width=\linewidth]{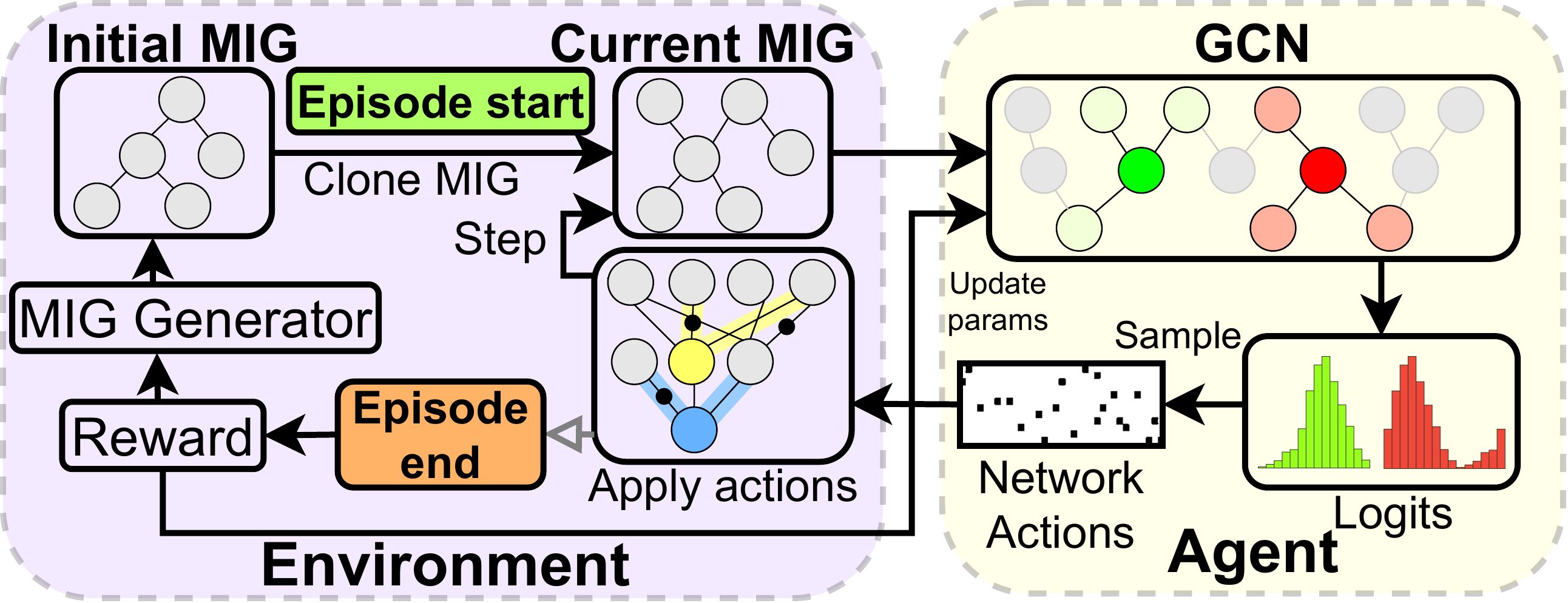}
\caption{Schema of the DRL system for logic optimization.}
\label{fig:system}
\end{figure}  
Our method, shown in \cref{fig:system}, performs logic synthesis on circuits composed by 3-input majority and inverter gates represented in our model as MIGs, whose algebra allows defining a sound and complete move set \cite{Amaru2016}. At every time step, the system applies the instructions provided by an RL agent which learns autonomously which transformations are best to apply to each circuit configuration. The training process is organized by episodes with a fixed number of steps. At episode start, an initial graph is generated and a sequence with sets of transformations is applied until episode end. The RL agent learns its policy by interacting with the \textit{environment}. This is responsible for consistently applying the agent's chosen legal transformations while blocking illegal transformation attempts. The latter includes handling collisions of simultaneous actions from distinct nodes. To help the agent explore the space more efficiently, we provide additional inference rules which ease applicability, mitigating the reward sparsity problem. In this setup, it is no longer needed to commute each node inputs in specific ways to allow action application. The \textit{environment} computes and sends to the agent the reward (size difference between initial and final graphs) achieved in an episode. Thus, all actions taken at episodes with size reduction are positively reinforced.

For a given node $n_{i}$ of a MIG and a maximum adjacency depth $d_{adj}$, we define the neighborhood $N_{i}$ of $n_{i}$ as the set of nodes $n_{j}$ which are reachable from $n_{i}$ traversing no more than $d_{adj}$ edges. We define a state space $S$ where $s\in S$ represents any subgraph that can be observed by the GCN in $N_{i}$. We call \emph{action} any graph transformation that keeps intact the truth table of the MIG (legal move). These actions are selected by the agent ($\Omega$ actions), or are always applied after each step ($\Lambda$ actions), cf. \cref{tab:graphTransforms}. Additionally, the identity action $\Omega \cdot I$ is always applicable. 

\begin{table}
    \centering
    \caption{Local Majority-Inverter Graph Transformations}
    \label{tab:graphTransforms}
    \resizebox{\linewidth}{!}{
    \begin{tabular}{@{}l@{\hspace{1mm}}l@{\hspace{1mm}}l@{}}
        \toprule
        $\Omega \cdot C$ & Commutativity & $M(x,y,z) = M(y,x,z) = M(z,y,x)$ \\
        $\Omega \cdot A$ & Associativity & $M(x,u,M(y,u,z) = M(z,u,M(y,u,x))$ \\
        $\Omega \cdot A_{c}$ & Compl. Assoc. & $M(x,u,M(y,u',z) = M(x,u,M(y,x,z))$ \\
        $\Omega \cdot D$ & Distributivity & $M(x,y,M(u,v,z) = M(M(x,y,z),M(x,y,v),z)$ \\
        $\Omega \cdot I$ & Inverter Prop. & $M'(x,y,z) = M(x',y',z')$ \\
        \midrule
        $\Lambda \cdot M$ & Majority & $M(x,x,z) = x$ and $M(x,x',z) = z$ \\
        $\Lambda \cdot R$ & Redundancy & 
        $\forall u, v \in $ nodes: if in($u$)=in($v$), replace all $u$ with $v$\\
        \bottomrule
    \end{tabular}
    }
\end{table}

At every step of an episode, the agent observes the state of each node $n_{i}$ in the MIG by performing a GCN forward pass. At each hidden layer, the network collects for each $n_{i}$ features vectors (FVs) from adjacent nodes $n_{k}$. The FVs of its 3 inputs are concatenated with 3 aggregated FVs for the output. The nodes connected to the output are binned based on its input port (0, 1, or 2) of the connection, then the FVs in each of the bins are summed up to produce the aggregated FVs. The resulting FVs are passed through linear layers and non-linearities, while the process is repeated for every layer. In this layout, the network explores for each $n_{i}$ a broader $N_{i}$ at each layer. Finally, the network outputs a distribution with action probabilities for each node given its state $P(a\mid s)$, and the agent samples an action for each node. In accordance, the environment applies the legal selected moves and this process is repeated until the episode ends, when the reward is computed and used to update the model parameters. We train our model with Policy Gradient, as the sharp changes on estimation policy require the explicit state exploration from on-policy algorithms. They also break ties on equivalent sequences by exploiting one of them. 

\begin{figure}
\centering
\includegraphics[width=0.97\linewidth]{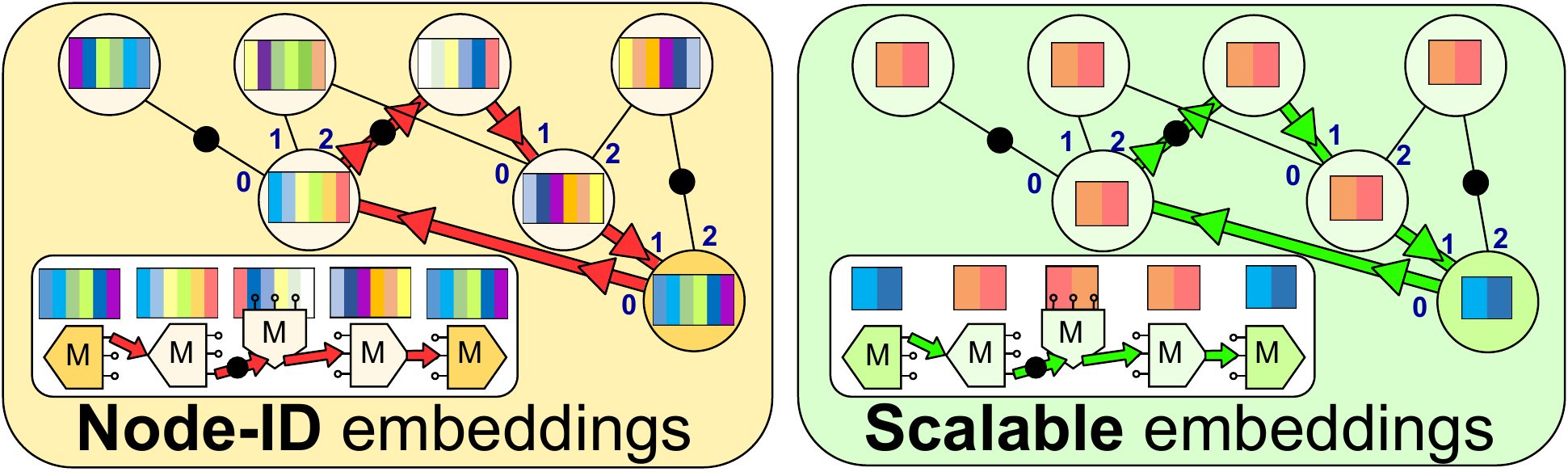}
\caption{Representation of the pattern seen by the model within a node neighborhood for (1) traditional \textit{Node-ID} and (2) our proposed \textit{scalable} embeddings. For (1), the model explores neighborhoods by comparing feature vectors through paths in all directions, while the latter unravels sequences of edges through closed loops of length up to $d_{adj}$. Below, the unraveled gathered sequences of features are shown, which encode the inverters and gates traversed by the path. }
\label{fig:embeddings}
\end{figure} 

\textbf{Model embeddings:} A major obstacle of GCN-based models is the need to embed each node with the necessary information. Common options are learned embeddings based on the node type, which lacks relevant information, and node IDs, which are non-scalable as they depend on the graph size.  
We propose an alternative, scalable embedding method in which every node can just ``see'' whether a node in its neighborhood is ``itself'' or ``not itself''. This implies that every node neighborhood is initialized with its node-specific embedding before a forward pass. In the following, we demonstrate that this is sufficient to learn the required features for logic optimization. \cref{fig:embeddings} depicts the feature collection process.

The proposed method nevertheless has some limitations. 1) The multi-action schema makes it impractical to use actions involving an arbitrary number of nodes, such as substitution or relevance. These have been proven to be effective in heuristic optimization approaches. However, the system can still get to any state without these actions in a few more steps \cite{Amaru2016}. 2) The specific embeddings for each node have a memory and compute time footprint depending on the graph's connectivity. However, this is not a scalability issue as it does not depend on graph size.

\section{Results}

 We train our model with 3-input and 4-input decompositions obtained through sum-of-products (SOP) and compare the mean size reduction obtained with the maximum achievable, computed with \textit{cirkit}'s exact synthesis \cite{Soeken}. We also train with two sets of 1000 random graphs with 100 inputs, 2 outputs and sizes $(50, 500)$. Finally, we train our model with \textit{C1355} and \textit{C880} benchmarks \cite{Iscas85}. Size reductions achieved with Random graphs and benchmarks are compared with the ones obtained with ABC \cite{Epfl}. The following table shows the average size reduction achieved with each dataset and method, and their relative improvement with respect to the baseline.

\begin{table}[h!]
\label{tab:meansizered}
\centering

\resizebox{0.9\linewidth}{!}{
\begin{tabular}{ lrrrrr }
\toprule
Dataset & baseline$^a$ & \multicolumn{2}{c}{------ DLLOA \cite{Haaswijk2018} ------} & \multicolumn{2}{c}{------ \textbf{ours} ------} \\
 & MSR$^b$ & MSR$^b$ & relMSR$^c$ & MSR$^b$ & relMSR$^c$\\
\midrule
SOP$_{3}$  & 8.65 & \textbf{8.65} & \textbf{100\%} & 7.38 & 85\% \\
SOP$_{4}$  & 24.59 & 20.41 & 83\% & \textbf{24.57} & \textbf{100\%}\\
Rand 50 & 25.54 & -- & -- & \textbf{44.92} & \textbf{175\%} \\
Rand 500 & 271.24 & -- & -- & \textbf{413.68} & \textbf{152\%} \\
C880\cite{Iscas85}  & \textbf{39} & -- & -- & 17 & 44\%\\
C1355\cite{Iscas85}  & 114 & 98 & 86\% & \textbf{106} & \textbf{93\%}\\
\bottomrule
\end{tabular}
}

\vspace{1mm}
\raggedright
\footnotesize{$^a$ SOP datasets: optimal synthesis (cirkit \cite{Soeken}); all others: ABC \cite{Epfl}.\\
$^b$ Mean Size Reduction: Average difference between initial and final graphs.\\
$^c$ Relative Mean Size Reduction: Average size reduction relative to baseline.\\
}
\end{table}

The model provides competitive results for the benchmarks and SOP datasets while outperforming ABC for randomly generated graphs. This difference in performance is influenced by the fact that ABC operates with AIGs, which resembles the original structure of benchmarks and SOPs, while our LO operates with MIGs, used for generating random graphs.

\section{Conclusion}
\label{sec:conclusion}
We have proposed to combine GCNs with RL with a novel node embedding to perform logic optimization on MIGs. We show promising results on several benchmark graphs, outperforming ABC with a size reduction of up to 1.75$\times$ on larger, random graphs.

\bibliographystyle{IEEEtran}
\bibliography{references}

\end{document}